# Global Context for improving recognition of Online Handwritten Mathematical Expressions

Cuong Tuan Nguyen[1][0000-0003-2556-9191], Thanh-Nghia Truong[1][0000-0002-8635-8534], Hung Tuan Nguyen[2][0000-0003-4751-1302], and Masaki Nakagawa[1][0000-0001-7872-156X]

[1] Department of Computer and Information Sciences, Tokyo University of Agriculture and Technology, 2-24-16 Naka-cho, Koganei-shi, Tokyo, 184-8588 Japan
[2] Institute of Global Innovation Research, Tokyo University of Agriculture and Technology, 2-24-16 Naka-cho, Koganei-shi, Tokyo, 184-8588 Japan
fx4102@go.tuat.ac.jp,{thanhnghiadk, ntuanhung}@gmail.com, nakagawa@cc.tuat.ac.jp

**Abstract.** This paper presents a temporal classification method for all three subtasks of symbol segmentation, symbol recognition and relation classification in online handwritten mathematical expressions (HMEs). The classification model is trained by multiple paths of symbols and spatial relations derived from the Symbol Relation Tree (SRT) representation of HMEs. The method benefits from global context of a deep bidirectional Long Short-term Memory network, which learns the temporal classification directly from online handwriting by the Connectionist Temporal Classification loss. To recognize an online HME, a symbol-level parse tree with Context-Free Grammar is constructed, where symbols and spatial relations are obtained from the temporal classification results. We show the effectiveness of the proposed method on the two latest CROHME datasets.

**Keywords:** Temporal Classification, Online Handwritten Mathematical Expression, Context-Free Grammar.

## 1    Introduction

Recognizing an online handwritten mathematical expression (HME) is a task of transcribing a sequence of pen-traces as coordination points input into a LaTeX or MathML description. It provides a simple and efficient user interface for mathematical input, where users can write mathematical expressions directly on a tablet-PC or by a digital pen. The user interface provides benefits for e-learning applications. Moreover, HME recognition is also useful for automatic marking [1] and intelligent tutoring system [2].

Generally, a stroke level parse tree and Context-Free Grammar (CFG) are used to represent the two-dimensional structure of an HME [3, 4]. An HME is recognized by constructing a parse tree from the results of three subtasks: symbol segmentation, symbol recognition, and relation classification. Due to separated models for the three subtasks, however, global context is not utilized for all subtasks. Besides, these models are highly dependent on handcrafted features.

An encoder-decoder model using Deep Neural Networks is one of the solutions for HME recognition, which performs all the subtasks in a single model [5]. The model benefits from the global context by deep neural networks and avoids the dependency of handcrafted features. However, the Deep Neural Network approach currently suffers from a high cost of recognition due to its complexity. Another problem is the challenge of utilizing grammar for improving performance.

The global context could be obtained through sequential models. In handwritten text recognition, temporal recognition of characters benefits from the bidirectional context of preceding strokes and succeeding strokes by Recurrent Neural Networks (RNNs) [6]. Nguyen et al. applied a bidirectional RNN model for improving symbol segmentation and recognition of online HMEs [7]. Nonetheless, classification of spatial relations (relation classification) between symbols was separated from other subtasks, which does not benefit from the global context. Zhang et al. applied a tree-based bidirectional RNN model to learn symbol recognition and relation classification from HMEs [8]. However, the approach exhibited difficulty to learn the relation classification.

In this work, we improve the performance of the three subtasks by introducing a single recognition model for them. This model uses a bidirectional RNN, which utilizes the global context for symbol recognition and relation classification. We apply a constraint to make the model output the relation classification at a precise time step to solve both symbol segmentation and relation classification. The model is trained using raw online HMEs so that it does not depend on handcrafted features. We apply a new parsing method at the symbol level instead of the stroke level to produce the result of HME recognition. In the parsing process, missing relations are reevaluated by applying the temporal classifier.

The rest of the paper is as follows: Section 2 introduces the related works, Section 3 describes our methods, Section 4 shows the experiment results and discussion, Section 5 draws our conclusion.

## 2 Related works

### 2.1 Tree-based BLSTM for HME recognition

Zhang et al. used a Stroke Label Graph (SLG) to represent an HME, in which nodes represent strokes and edges represent relations between the strokes. The graph is represented by a bidirectional Long Short-term Memory (BLSTM) or by a tree-structured BLSTM (tree-BLSTM) [8, 9]. These models took advantage of the bidirectional context by BLSTM to learn the classification directly from raw online features so that they achieved high symbol classification rates. However, the models are still troubled with learning spatial relations. Due to lack of coverage on spatial relations learned by sequential input or tree-based input, multiple BLSTM models or tree-BLSTM models were used to learn different sequences or tree derivations from the SLGs.

Tree-BLSTM firstly suffers from learning by stroke-level constraint, where the spatial relations from strokes are complex. Second, it encounters difficulty in integrating separated tree models. Consequently, there is a significant drop in the recall rate of relation classification [8, 9].

### 2.2 Spatial relations classification

For recognizing spatial relations, the common approach depends on handcrafted features such as geometrical descriptors [3], shape descriptors [10]. The handcrafted features are sensitive to ambiguous spatial relations since they are extracted from bounding boxes of symbols [3] or their centroids [4]. To deal with the problem, they modified the descriptors to add the dependency on symbol classes. The symbol classes were grouped into four groups of Ascendant, Descendant, Normal, and Big. Then, bounding boxes were modified as "body box" while vertical centers were shifted according to each group [3, 4]. These modifications also need to be carefully designed and may not be robust for various handwriting styles. Despite the modifications, the approach is still problematic with the *Subscript* relation due to its confusion to the *Right* relation.

The BLSTM and tree- BLSTM models could learn spatial relations directly from features [8, 9]. However, they perform a limited accuracy and have a problem with the *Subscript* relation.

### 2.3 Symbol recognition with Bidirectional Context

Nguyen et al. applied a temporal classification method for recognizing mathematical symbols [7]. The method uses a BLSTM model to take advantage of the bi-directional context for classifying symbols. The method improves symbol classification rate and expression recognition rate of the CFG based parse tree for online HME recognition. The results showed the effectiveness of applying global context to symbol recognition. The segmentation of the online HME recognition system is improved by "junk detection," and it also benefits from the temporal classifier with the global context. For spatial relations classification, however, the method is still relying on geometrical descriptors.

## 3 Our method

In this section, we propose a single model for learning all the three subtasks of online HME recognition and its integration to a parse tree for recognizing the whole HME.

### 3.1 Overview of the proposed method

Our method consists of two stages, as shown in Fig. 1. A sequence of extracted features from an online HME input is fed to a deep BLSTM symbol-relation temporal classifier to predict symbols and relations in the first stage. In the second stage, the symbols and relations from the temporal classifier are input to a symbol-level parser to build the parse tree and the recognition result is output as a Latex string.

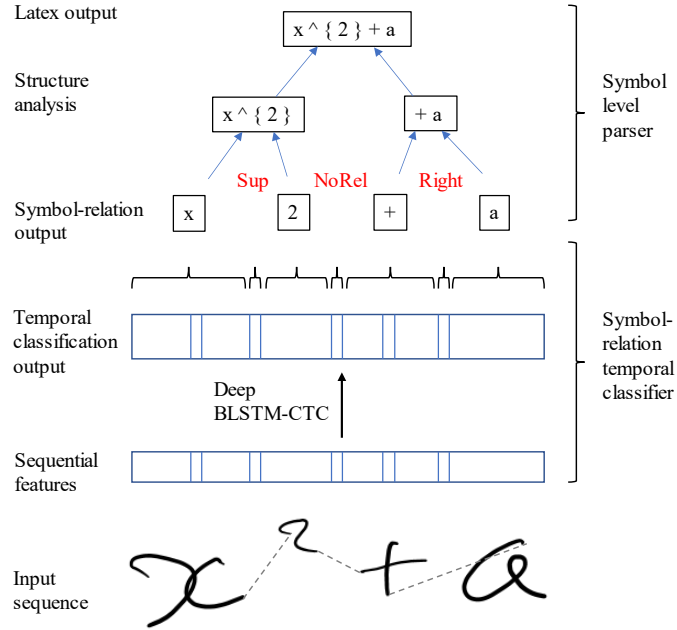

**Fig. 1.** Overview of the proposed method.

### 3.2 Sequential model for segmentation, recognition, and relation classification

The symbol relation tree (SRT) or symbol layout tree [11] is an effective way to represent the structure of an HME, where symbols are represented as nodes and spatial relations between symbols are represented as links between nodes [12]. Fig. 2 shows an example of an input online HME and its SRT representation in Fig. 2(a) and Fig. 2(b), respectively. Each node contains a symbol label and indexes of the strokes that belong to the symbol. A link between two nodes is a directed connection that denotes a spatial relation from the parent node to the child node. There are six types of spatial relations in SRT: *Above*, *Below*, *Sub* (subscript), *Sup* (superscript), *Right*, and *Inside* (square root).

The whole structure of an HME could be represented by derived paths of consecutive symbols and spatial relations between them from its SRT. Fig. 2 (c) shows derived paths by tracing from the root to all leaves of the SRT. Each derived path contains the information of the input sequence and its label sequence, which can be learned by a sequential model.

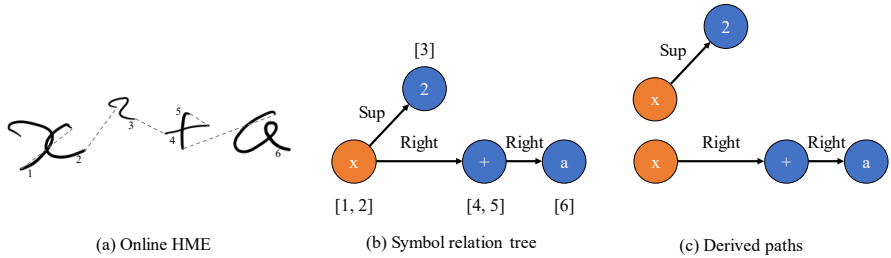

(a) Online HME     (b) Symbol relation tree     (c) Derived paths

**Fig. 2.** Symbol relation tree and its derived paths.

Here, we propose a sequential model for recognizing both symbols and spatial relations. The model is a temporal classifier that uses a feature sequence as input and output the sequence of symbols and spatial relations between two consecutive symbols. The temporal classifier produces the probabilities of symbols and relations for every time step of the input sequence. The input sequence consists of both strokes and off-strokes, where an off-stroke is a virtual pen trace between two consecutive strokes, connecting the end of the first stroke to the beginning of the second stroke.

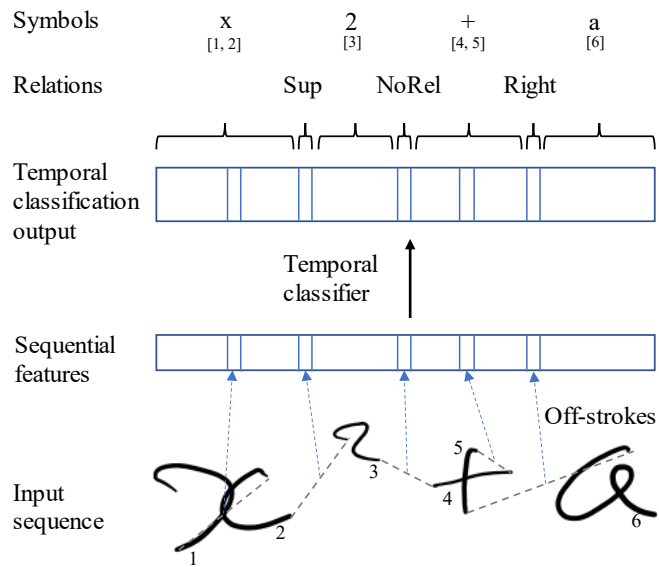

**Fig. 3.** Temporal classifier for symbol segmentation, symbol recognition, and relation classification.

Assuming the sequential order of input strokes and symbols are consistent (i.e., there are no delayed strokes), symbols are separated by off-strokes between them. We propose a temporal classifier that predicts symbols and spatial relations at the precise time steps of the corresponding strokes and off-strokes, as shown in Fig. 3. For off-strokes between two strokes inside a symbol, there is no output. From the output of the temporal classifier, we obtain both the segmentation positions and the spatial relations between symbols at the off-stroke locations. Then, the symbols are separated their classification are obtained. Therefore, we can solve three tasks of symbol segmentation, symbol recognition, and relation classification using the single temporal classifier.

### 3.3 Deep BLSTM for segmentation, recognition, and relation classification

We apply a deep BLSTM to incorporate bidirectional context for symbol and relation classification. The model is a stack of multiple BLSTM layers where each layer is a combination of two LSTM layers that process the input in forward and backward directions [13]. The forward and backward context by the two LSTM layers is combined and fed to the next BLSTM layer in the networks. The deep architecture is for extracting high-level features directly from the raw input sequence, while BLSTM benefits from the long-range context of forward and backward directions by LSTMs.

For training the deep BLSTM network, a Connectionist Temporal Classifier (CTC) [6] loss function is applied. CTC helps the network learn from the input sequence to a target sequence without explicit alignment or segmentation needed. The alignment between the input sequence and output label sequence is learned automatically with the assumption that the two sequences are in the same order.

CTC introduces a label called 'blank' that denotes no label. It defines the output $y^t$ of RNN for each time step $t$ with respect to an input sequence $x$ with the length $T$ as the probability distribution over a fixed set of classes $C$ and the 'blank' label as shown in Eq. (1).

$$y_k^t = p(k, t \mid x), \forall k \in C \cup blank \tag{1}$$

where $y_k^t$ is the output $y^t$ for a class $k$.

An output label sequence $l$ is obtained by a reduction process $B$ over a path $\pi_{1:T} = k_1, k_2, .., k_T$ through the lattice of output labels, i.e. $k_i \in C \cup blank, i = \overline{1, T}$. The reduction process firstly removes repeated labels, then removes 'blank' labels in this path. The probability for an output label sequence $l$ from an input sequence $x$ is the total probability of all the paths $\pi_{1:T}$ such that each path $\pi_{1:T} \in B^{-1}(l)$ is reduced into $l$, as shown in Eq. (2).

$$p(l|x) = \sum_{\pi_{1:T} \in B^{-1}(l)} p(l, \pi_{1:T}|x) \tag{2}$$

where $p(l, \pi_{1:T}|x) = \prod_{t=1}^{T} p(k_t, t|x)$ is the probability of the label sequence $l$ over the path $\pi_{1:T}$.

For a pair of an input sequence $x$ and an output sequence $l$ from the training dataset, the network is trained by minimizing the CTC loss obtained by Eq. (3).

$$loss_{CTC} = -log\big(p(l|x)\big) \tag{3}$$

From the sequential model, segmentation is performed by finding off-strokes with high probabilities of relations. For an online HME, let $S$ is a sequence of $n$ strokes of the HME as $S = (s_0, \ldots, s_{n-1})$ and $O$ is a sequence of $(n-1)$ off-strokes as $O = (o_1, \ldots, o_{n-1})$ where $o_i$ is an off-stroke between two strokes $s_{i-1}$ and $s_i$, $\varphi_{HME}$ is the BLSTM context when parsing $S$ with the symbol-relation temporal classifier. The $i^{th}$ off-stroke $o_i$ can be a relation or 'blank' character between two strokes inside a symbol. The relation between the $(i-1)^{th}$ symbol and the $i^{th}$ symbol is obtained from the relation probability at the $i^{th}$ off-stroke, which is calculated as shown in Eq. (4):

$$Rel(o_i) = \begin{cases} argmax \left( P_{rel}(o_i | \varphi_{HME}) \right) if \ max \left( P_{rel}(o_i | \varphi_{HME}) \right) \geq P_{blank} \\ 'blank' if max \left( P_{rel}(o_i | \varphi_{HME}) \right) < P_{blank} \end{cases} \quad (4)$$

where:

- $Rel(o_i)$ is the predicted relation of $i^{th}$ off-stroke.
- $P_{rel}(o_i | \varphi_{HME})$ is the relation probability of the relation "rel" at $o_i$.
- $P_{blank}$ is the probability of $o_i$ being a 'blank' character.

Symbol recognition is performed by taking the maximum probability of symbols between two relation outputs. The symbol recognition for a list of $t$ consecutive strokes $(s_i, \ldots, s_{i+t})$ is computed as shown in Eq. (5):

$$Symbol(s_{i:i+t}) = argmax \left( P_{symbol}(s_{i:i+t} | \varphi_{HME}) \right) \quad (5)$$

where $P_{symbol}(s_{i:i+t} | \varphi_{HME})$ is the probability of symbol recognition from stroke $s_i$ to stroke $s_{i+t}$.

### 3.4 Constraint for output at precise time steps

Zhang et al. used the "local CTC" learning method, which applied a constraint on the output of symbols and relations at the specific strokes and off-strokes [8]. The method, however, does not constrain the output of relations to the exact time steps so that it causes an ambiguity for learning relation classification on many possible time steps.

As BLSTMs could learn precise time steps [14], we propose a single feature point for representing each off-stroke and applying the constraint loss shown in Eq. (6). The constraint loss is in the form of a binary cross-entropy loss, which penalizes the relations output at the strokes. Thus, the model is forced to learn to output relation classification at the off-strokes.

$$loss_{CE} = -\sum_{i=0}^{n-1} log \left( 1 - \sum P_{rel}(s_i | \varphi_{HME}) \right) \quad (6)$$

The deep BLSTM model is trained using a combination of the constraint loss and CTC loss as shown in Eq. (7).

$$loss = loss_{CTC} + \lambda loss_{CE} \quad (7)$$

where $\lambda$ is a weighted parameter that is determined experimentally.

### 3.5 Training path extractions

From an SRT, we extract multiple derived paths of input stroke sequences from the stroke indexes of the symbol sequence for training the temporal classifier. Each of them represents a path of strokes and off-strokes, as well as their symbol labels and the spatial relations among them.

We propose three methods of path extractions:

1. Trace all paths from the root to all leaves of an SRT
2. Trace the path by writing order. *NoRel* is added when there is no relation between two consecutive nodes.
3. Extract random paths. Algorithm 1 comprehensively presents how we randomly extract derived paths. The method randomly shuffles the order of sub-trees connected to the parent node and then combines them to simulates various writing orders.

---

**Algorithm 1**: ExtractRandomPath

**Input:**
    Root node: $r$.
**Output:**
    List of nodes as a path: $p$.

| | |
|---|---|
| 1 | **if** $|r.\text{childs}| = 0$ **then** |
| 2 |   $p \leftarrow [r]$ |
| 3 | **else if** $|r.\text{childs}| = 1$ **then** |
| 4 |   $p \leftarrow [r, \text{ExtractRandomPath}(r.\text{childs}[0])]$ |
| 5 | **else** |
| 6 |   $p = [\quad]$ |
| 7 |   **for** $node$ in shuffle $([r, r.\text{childs}])$ **do** |
| 8 |     **if** node $= r$ **then** |
| 9 |       $p \leftarrow [p, node]$ |
| 10 |     **else** |
| 11 |       $p \leftarrow [p, \text{ExtractRandomPath}(node)]$ |
| 12 |     **end if** |
| 13 |   **end for** |
| 14 | **end if** |
| 15 | **return** $p$ |

---

Fig. 4 and Fig. 5 show examples of how we extract derived paths from an SRT. Fig. 4 (a) presents the example of tracing all derived paths from the root node of the SRT. In this case, all the spatial relations can be extracted except *NoRel*. Fig. 4(b) presents an example of tracing a derived path by the writing order, which may contain *NoRel*. To extract more *NoRel* relations, we propose the third path extraction methods. Fig. 5 shows an example of generating derived paths by randomly shuffling child nodes of the SRT.

The path extraction methods generate multiple derived paths from a single SRT. Therefore, our proposed method can generate a large number of patterns for training the temporal classifier.

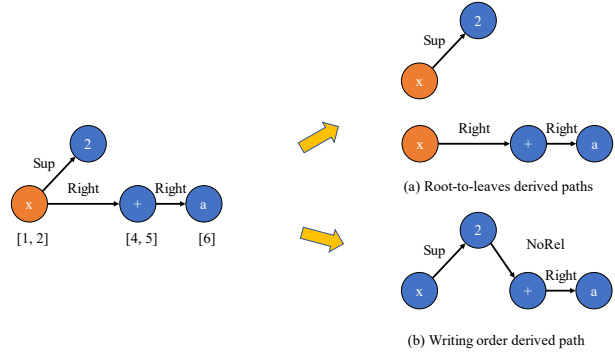

(a) Root-to-leaves derived paths

(b) Writing order derived path

**Fig. 4.** Path derived from SRT.

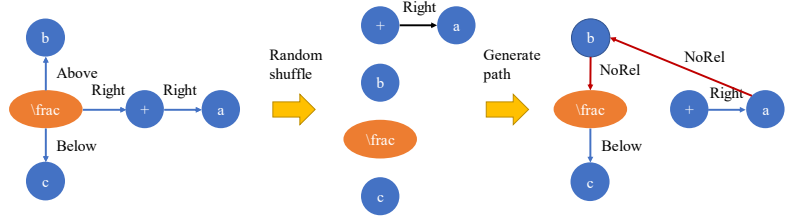

**Fig. 5.** Random path generation by shuffling child nodes.

### 3.6 Symbol level parse tree

**Symbol level syntactic parse tree.** We use a two-dimensional Context-Free Grammar (2D-CFG) G of a 4-tuple $(\mathcal{N}, \Sigma, S, P)$, where $\mathcal{N}$ is a finite set of nonterminal symbols, $\Sigma$ is a finite set of terminal symbols, $S \in \mathcal{N}$ is the start symbols of the grammar, and $P$ is a finite set of rules in Chomsky Normal Form (CNF) as in the form of terminal rules $A \rightarrow a$ and binary rules $A \xrightarrow{r} BC$, with $A, B, C \in \mathcal{N}$, $a \in \Sigma$, and $r$ denoting the spatial relation between $B$ and $C$.

To generate the parse tree from the symbol sequence obtained from the temporal classifier, we use the bottom-up Cocke-Younger-Kasami (CYK) parsing algorithm. The probability for a terminal production $A \rightarrow a$ and that for a binary production $A \xrightarrow{r} BC$ are obtained by Eq. (8) and Eq. (9), respectively.

$$P(A) = P(a) \tag{8}$$

$$P(A) = P(B) \times P(C) \times P(r) \tag{9}$$

where $P(B)$, $P(C)$ are obtained by the previous production recursively, $P(a)$ is the probability of terminal symbols obtained from the temporal classifier, $P(r)$ is the prob-ability of spatial relation between two nonterminal symbols B and C that could be derived from spatial relation between two terminal symbols by the temporal classifier. If a spatial relation between two terminal symbols is not available from the temporal sequential input, we build a sequential input through the two terminal symbols and apply the temporal classification to obtain the spatial relation.

Since the temporal classifier provides many recognition candidates for symbols and relations, CYK parsing generates HME recognition candidates at the root of the parse tree with their production probabilities. We obtain the HME recognition result by selecting the candidate of the highest probability at the root of the parse tree.

**Spatial relations for nonterminal symbols.** To obtain the probability $P(r)$ of a spatial relation between two nonterminal symbols, we need to determine the two terminal symbols representing that relation and obtain the probability from the temporal classifier. Here, we make a rule to obtain the spatial relation between two non-terminal symbols by the relation between the two terminal symbols inside them.

First, we divide the terminal symbols into two types of:

— Dominant symbols (\sqrt, \frac, \lim, \sum, \int)
— Non-dominant symbols: the remaining symbols

Then, we define the following terms:

— A component of a nonterminal symbol is a terminal symbol or a group of terminal symbols that belong to the nonterminal symbol.
— Baseline of a nonterminal symbol is a list of terminal symbols extended from the left most symbol by the *Right* relation.
— Six types of components in a nonterminal symbol: Main, Left, Right, Left-right, R-Sup, R-Sub. The Left component is the left most terminal symbol. The Right component is the right most terminal symbol in the baseline. The Left-right component is composed of Left and Right. R-Sup is the last terminal symbol extended from a Right component by the *Sup* relation. The same definition is applied for R-Sub by the *Sub* relation. The Main component is a dominant symbol in the baseline when there is no other symbol in the baseline.

For a nonterminal symbol containing the Main component, all the inbound and outbound relations are applied to the Main component. Otherwise, the inbound and outbound relations are connected by the rules presented in Fig. 6.

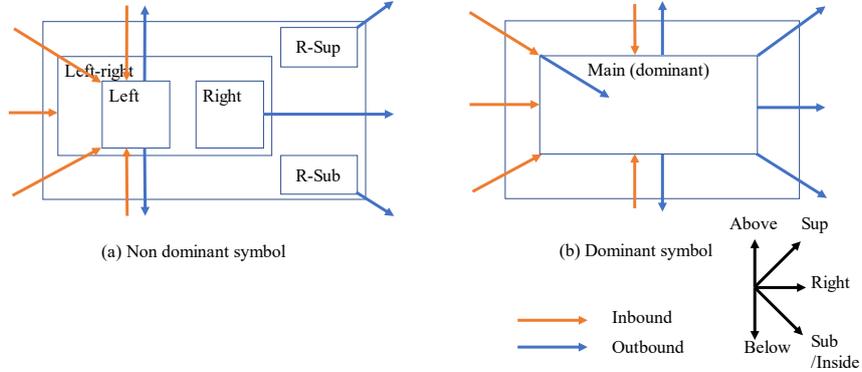

**Fig. 6.** Spatial relation rules for nonterminal symbols.

## 4    Evaluation

### 4.1    Dataset

We conducted the experiments on the CROHME competitions dataset [12] for both the three subtasks and HME recognition. We used only the official CROHME training set for training the temporal classifier. The number of classes for symbols is 101 and that for spatial relations is 7 composed of {*Above*, *Below*, *Sub*, *Sup*, *Right*, *Inside*, *NoRel*}.

### 4.2    Setup for experiments

We use a stack of three BLSTM layers. Each BLSTM contains two LSTM layers with 128 cells. The outputs of each time step by two LSTM layers are concatenated into a feature vector of 256 dimensions before input into the next BLSTM layer. We train the networks using Stochastic Gradient Descent (SGD) with a learning rate of 0.0001 and a momentum of 0.9. The $\lambda$ parameter of the combined loss function is set to 0.1.

In our experiment, an input sequence of coordinates in a stroke were sampled by the Ramer method [15]. For each sampled point, we then extracted four features: the sine and cosine of the writing directions, the normalized distance between the preceding and the succeeding points of the current point, and a binary value of pen state (pen-up/pen-down).

We measured the performance of symbol segmentation, symbol recognition, and relation classification of the model through recognizing online HMEs. The expression rate was evaluated by the provided Symbolic Label Graph (SymLGs) using the LgEval tool [16].

### 4.3    Experiments and results

**Subtasks evaluation**. The proposed model is firstly evaluated on the three subtasks of symbol segmentation, symbol recognition, and relation classification. For the evaluation, we obtained the prediction results from the temporal classifier as symbols and relations. Symbol segmentation and symbol recognition were evaluated on the sequential input of online HMEs in their writing order. For evaluating relation classification, we prepared 10 random sequential inputs derived from each SRT for better coverage of spatial relations.

Table 1. shows the results of the three subtasks of our method compared with the other state-of-the-art methods. As compared with the methods using separated modules of the three subtasks, our method shows the improvement in a large margin and gets a competitive result as MyScript, which used an extra dataset for training. As compared with the SRT learning approach of BLSTM or tree-BLSTM, our method shows a considerable improvement, especially on relation classification.

**Table 1.** Symbol level evaluation on CROHME 2016 testing set.

| Type | System | Seg. (%) | | Seg. + Cls. (%) | | Tree rel. (%) | |
|---|---|---|---|---|---|---|---|
| | | Rec. | Prec. | Rec. | Prec. | Rec. | Prec. |
| Separated modules | MyScript | 98.89 | 98.95 | 95.47 | 95.53 | 95.11 | 95.11 |
| | Wiris | 96.49 | 97.09 | 90.75 | 91.31 | 90.17 | 90.79 |
| | Tokyo | 91.62 | 93.25 | 86.05 | 87.58 | 82.11 | 83.64 |
| | Sao Paulo | 92.91 | 95.01 | 86.31 | 88.26 | 81.48 | 84.16 |
| | Nantes | 94.45 | 89.29 | 87.19 | 82.42 | 73.20 | 68.72 |
| Single module | BLSTM [9] | 92.77 | 85.99 | 85.17 | 78.95 | 67.79 | 67.33 |
| | Tree-BLSTM [8] | 95.52 | 91.31 | 89.55 | 85.60 | 78.08 | 74.64 |
| | Our system | **97.79** | **98.14** | **91.96** | **92.30** | **94.54** | **94.70** |

**Table 2.** Confusion rate matrix for relation classification on CROHME 2016 (%).

| Predict G.Truth | Above | Be-low | In-side | Right | Sub | Sup | NoRel | Non-Seg | Total |
|---|---|---|---|---|---|---|---|---|---|
| Above | **98.94** | 0.00 | 0.07 | 0.00 | 0.00 | 0.00 | 0.59 | 0.40 | 2520 |
| Below | 0.00 | **94.83** | 0.00 | 0.52 | 0.00 | 0.00 | 1.47 | 3.18 | 2898 |
| Inside | 0.05 | 0.69 | **93.07** | 0.75 | 0.00 | 1.44 | 2.45 | 1.55 | 1569 |
| Right | 0.02 | 0.01 | 0.01 | **94.55** | 0.93 | 0.22 | 3.14 | 1.13 | 64733 |
| Sub | 0.00 | 0.28 | 0.02 | 9.00 | **87.91** | 0.08 | 1.62 | 1.08 | 5816 |
| Sup | 0.50 | 0.00 | 0.00 | 2.01 | 0.75 | **92.84** | 3.70 | 0.20 | 9341 |
| NoRel | 0.09 | 0.11 | 0.03 | 3.53 | 0.08 | 0.35 | **95.47** | 0.33 | 60493 |
| NonSeg | 0.04 | 0.01 | 0.02 | 0.91 | 0.07 | 0.04 | 0.56 | **98.35** | 54310 |

Table 2 shows the confusion matrix of relation classification on the CROHME 2016 testing set. The worst case is the *Sub* relation of 87.91% due to the confusion to the Right relation. For other relations, the classification rate is from 92.84% to 98.94%.

Our model shows higher robustness with the confusion of recognizing the *Sub* relation compared with the other methods using the geometric and shape descriptors without context [10] or tree-BLSTM [8].

**Expression evaluation**. Our system produced a Latex sequence for each testing sample, then we converted it into SymLGs and used the LgEval tool to obtain the expression rate and structure rate.

Table 3 shows the expression rate of the method with comparing with the other method on CROHME 2016 and CROHME 2019 testing set. We omitted the results of some teams that using offline HME recognition methods. On CROHME 2016, our system achieved 53.44%, which is more than 3 points higher compared with Wiris, and more than 9 points higher compared with BLSTM_CTC [7]. This effect is from the improvement of all three subtasks. Our method also improves the expression recognition rate in a large margin compared with tree-BLSTM and LSTM of Nantes. Our method outperforms TAP* [5] of four models ensembled with more than 3 points. The result shows the effectiveness of our model in using global context over the deep learning approach. On CROHME 2019, our method achieved 52.38%, which is a competitive result for a single model without applying a language model, ensemble models and combination of online and offline recognition models as in TAP*+WAP*+LM / USTC system. The top systems of companies: MyScript, Samsung, MathType, also benefit from using their own extra dataset for training the system. As compared with BLSTM_CTC, our single model without a language model achieves an expression rate of more than 10 points higher.

**Table 3.** Expression rate (%) of the state-of-the-art methods.

| System | CROHME 2016 | | CROHME 2019 | |
|---|---|---|---|---|
| | Correct | Structure | Correct | Structure |
| MyScript | 67.65 | 88.14 | 79.15 | 90.66 |
| Wiris / MathType | 49.61 | 74.28 | 60.13 | 79.15 |
| Tokyo | 43.94 | 61.55 | 39.95 | 58.22 |
| Sao Paulo | 33.39 | 57.02 | - | - |
| Nantes | 13.34 | 21.45 | - | - |
| Samsung R&D [17] | 65.76 | - | 79.82 | 89.32 |
| TAP* [5] | 50.22 | - | - | - |
| TAP*+WAP*+LM [5] / USTC | 57.02 | - | **80.73** | **91.49** |
| Tree-BLSTM* [8] | 27.03 | - | - | - |
| BLSTM_CTC [7] | 44.81 | 60.94 | 41.28 | 56.80 |
| Our system | 53.44 | 76.02 | 52.38 | 72.64 |

\* denotes an ensemble of models

**Error Analysis.** We show some samples in Fig. 7, where (a), (b), and (c) are common errors in online HME recognition. Fig.7(a) presents a misrecognized symbol 'f' and a wrong relation with the following symbols of 'f', caused by a wrong writing order of the small bar stroke in the second symbol 'f'. Fig. 7(b) shows another misrecognized symbol due to the cursive writing style. Fig. 7(c) has all correctly recognized symbols, but there is a misclassified spatial relation between '\beta' and 'a' symbols. Finally, Fig. 7(d) shows a correctly recognized sample in both symbol and relation aspects, where the HME is written in a slope up direction. This shows the robustness of our method in relation classification.

Fig. 8 presents the expression recognition rates concerning the number of symbols in an HME. For the short HMEs having at most 10 symbols, the recognition rates are higher than 60%. Meanwhile, the recognition rates significantly reduce to lower than 20% for the long HMEs with more than 20 symbols. For the other HMEs consisting of between 11 to 20 symbols, the recognition rates seem adequate but might be improved in the future.

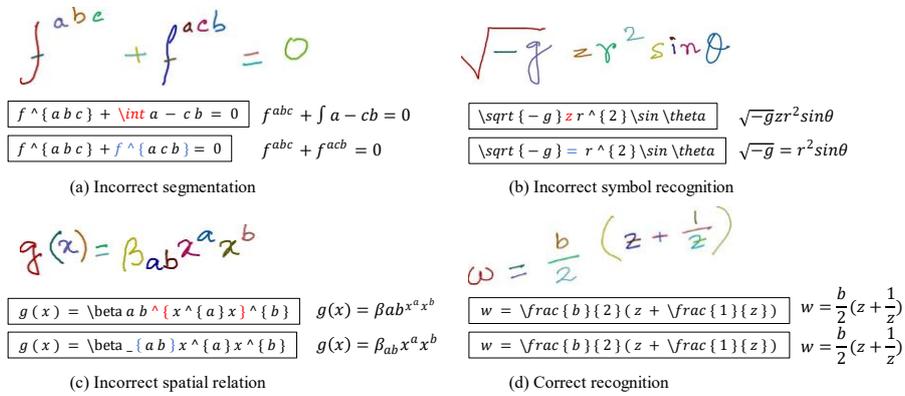

(a) Incorrect segmentation

(b) Incorrect symbol recognition

(c) Incorrect spatial relation

(d) Correct recognition

**Fig. 7.** Recognition results on Online HME samples.

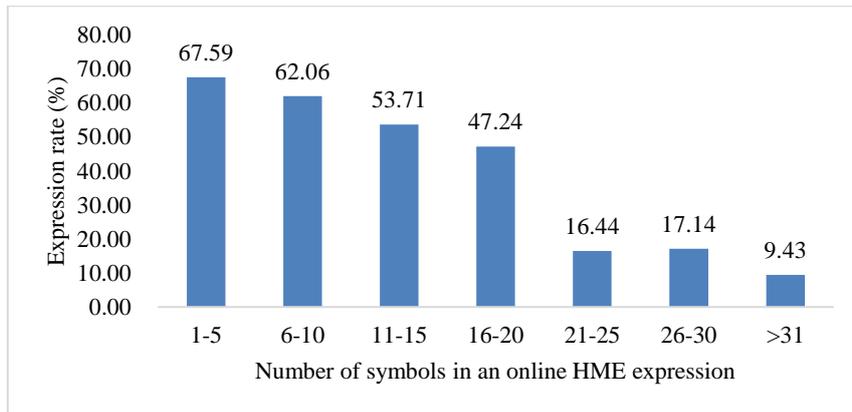

**Fig. 8.** Expression rate with respect to the number of symbols.

## 5    Conclusion

This paper proposed a temporal classification method for all three subtasks of symbol segmentation, symbol recognition, and spatial relation classification in online HME recognition. The method utilizes the global context of a bi-directional Long Short-term Memory network to improve the recognition rate of all subtasks. For recognizing a whole mathematical expression, a symbol-level parse tree built on top of the temporal classifier utilizes both the advantages of the temporal classifier and the Context-Free Grammar to improve the expression recognition rate. The proposed method achieves competitive results on both the CROHME 2016 and CROHME 2019 datasets by the single model without depending on a language model.

Further improvements of the approach are related to solving the problem of delayed handwriting, applying data augmentations and language models.

## Acknowledgement

This research is being partially supported by the grant-in-aid for scientific research (A) 19H01117 and that for Early Career Research 18K18068.